\documentclass[journal]{IEEEtran}

\ifCLASSINFOpdf
\else
\fi

\usepackage{times}  
\usepackage{helvet} 
\usepackage{courier}  
\usepackage[hyphens]{url}  
\usepackage{graphicx} 
\usepackage{caption} 

\usepackage{lineno}
\usepackage{url}            
\usepackage{booktabs}       
\usepackage{amsfonts}       
\usepackage{nicefrac}       
\usepackage{microtype}      

\usepackage{graphicx}
\usepackage{amsmath,amssymb} 
\usepackage{color}
\usepackage[ruled,vlined]{algorithm2e}

\usepackage{cleveref}
\usepackage{color,soul}

\renewcommand{\hl}[1]{#1}

\makeatletter
\def\endthebibliography{%
	\def\@noitemerr{\@latex@warning{Empty `thebibliography' environment}}%
	\endlist
}
\makeatother

\begin{document}
%
\title{Photorealism in Driving Simulations: Blending Generative Adversarial Image Synthesis with Rendering}
%
%
%

\author{Ekim~Yurtsever$^1$,~\IEEEmembership{Member}
        Dongfang~Yang$^{1,2}$,~\IEEEmembership{Student~Member}
        Ibrahim Mert Koc$^1$,~\IEEEmembership{Student~Member}
        and~Keith~A.~Redmill$^1$,~\IEEEmembership{Senior Member}
\thanks{$^1$ Department
of Electrical and Computer Engineering, the Ohio State University, Columbus,
OH, 43210 USA (email: yurtsever.2@osu.edu).}
\thanks{$^2 $ Chongqing Changan Automobile Co Ltd, Chongqing, Chongqing, CN}}

%
%

\markboth{~}%
{Shell \MakeLowercase{\textit{et al.}}: Bare Demo of IEEEtran.cls for IEEE Journals}
%



\maketitle

\begin{abstract}
Driving simulators play a large role in developing and testing new intelligent vehicle systems. \hl{The visual fidelity of the simulation} is critical for building vision-based algorithms and conducting human driver experiments. \hl{Low visual fidelity breaks immersion for human-in-the-loop driving experiments.} Conventional computer graphics pipelines use detailed 3D models, meshes, textures, and rendering engines to generate 2D images from 3D scenes. These processes are labor-intensive, and they do not generate photorealistic imagery.   Here we introduce a hybrid generative neural graphics pipeline for improving the visual fidelity of driving simulations. \hl{Given a 3D scene, we} partially-render only important objects of interest, such as vehicles, and use generative adversarial processes \hl{to synthesize} the background and the rest of the image. To this end, we propose a novel image formation strategy to form 2D semantic images from 3D scenery consisting of simple object models without textures. These semantic images are then converted into photorealistic RGB images with a state-of-the-art Generative Adversarial Network (GAN) trained on real-world driving scenes. \hl{This replaces repetitiveness with randomly generated but photorealistic surfaces.} Finally, the partially-rendered and GAN synthesized images are blended with a blending GAN. \hl{We show that the photorealism of images generated with the proposed method is more similar to real-world driving datasets such as Cityscapes and KITTI than conventional approaches. This comparison is made using semantic retention analysis and Frechet Inception Distance (FID) measurements.}

\end{abstract}


%
\IEEEpeerreviewmaketitle

\section{Introduction}
%
%
%
%
\IEEEPARstart{D}{riving} simulations are \hl{important} for developing and evaluating intelligent transportation systems \cite{punzo2010integration}. A good simulation environment should have accurate vehicle dynamics, realistic traffic behavior, and high visual fidelity. Visual fidelity is especially crucial for \hl{validating} vision-based algorithms and conducting \hl{human-in-the-loop} experiments. There are numerous studies \cite{lanata2014autonomic, ludl2020enhancing, minhas2020effects, 9295379, yang2019automated, ju2020acoustic} \hl{that utilize} a driving simulation whose integrity \hl{greatly depends} on the visual quality of the simulation environment.

The aforementioned studies all use \hl{rendered images that are} generated by a simulation \hl{environment}. However, limited work has been \hl{done on} evaluating \hl{and improving the visual fidelity of state-of-the-art driving simulators.} Here we investigate a new \hl{approach}: introducing \hl{generative} photorealism to virtual driving environments using deep learning. \hl{Data-centric applications trained or fine-tuned in a photorealistic driving simulation can be more confidently deployed to the real world.  Furthermore, automated driving systems can be tested with photorealistic-looking dangerous scenes that are difficult to obtain outside a simulation environment. In addition, if non-realistic repetitive patterns can be replaced by photorealistic scenery, the degree of immersion for human-in-the-loop simulation experiments can be increased.}

\begin{figure}[t!]
	\centering
	\includegraphics[width=1\linewidth]{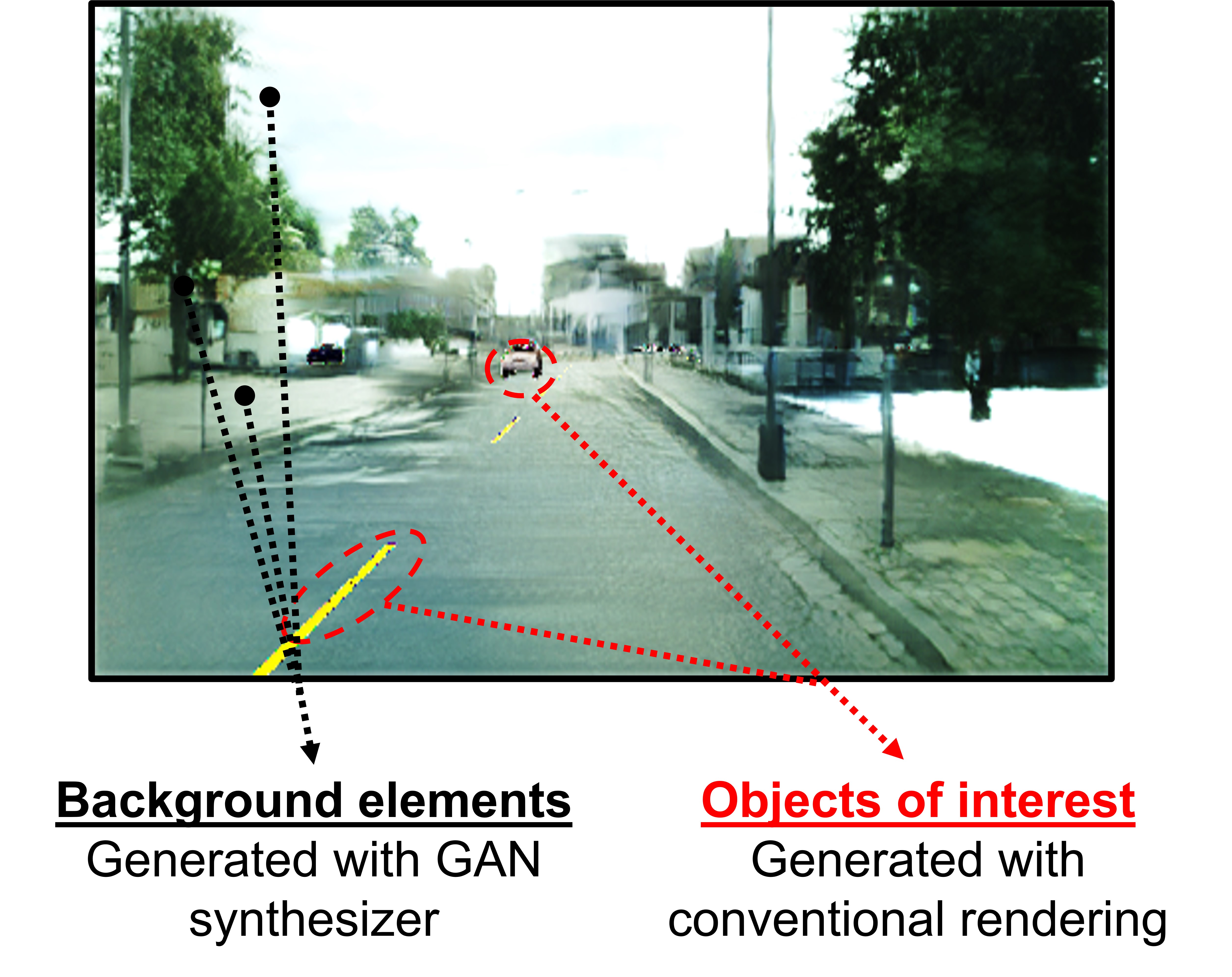}
	\caption{\hl{The proposed framework generates photorealistic imagery for driving simulators. First, we obtain the semantic layout of the scene through a conventional simulation pipeline with textureless simple 3D models. Then, this semantic layout is converted into a photorealistic RGB image using GANs with the proposed image formation and blending strategy.} } 
	\label{fig:sim_overview}
	\vspace{-5mm}
\end{figure}

\hl{The fidelity of a conventional driving simulator} depends on the quality of its computer graphics pipeline, which consists of 3D models, textures, and a rendering engine. High-quality 3D models and textures require artisanship, whereas the rendering engine must run complicated physics calculations for the realistic representation of lighting and shading \cite{kajiya1986rendering}. \hl{These processes are labor-intensive, and images obtained this way are not photorealistic.} Here we investigate alternatives for alleviating the aforementioned costs.  An overview of our approach is shown in Figure \ref{fig:sim_overview}.

The alternative to rendering is neural network based generative adversarial image synthesis. The advent of Generative Adversarial Networks (GAN) \cite{goodfellow2014generative} \hl{enabled} the realization of photo-realistic image synthesis \cite{brock2018large,zhang2018self,zhang2017stackgan,zhang2018stackgan++,bau2018gan,liu2017unsupervised,miyato2018cgans}.  A particular sub-problem, conditional image synthesis \cite{isola2017image,chen2017photographic,park2019semantic,qi2018semi, wang2018high}, delves into the more specific task of mapping a pixel-wise semantic layout to a complying photo-realistic image. The conditional semantic layout is the key link between the 3D scene and the generative synthesizer in our framework. More recently, video-to-video synthesis \cite{wang2018video} was proposed as an alternative to image synthesis. The temporal dimension was added to the generative process to reduce inconsistencies between synthesized frames.

\hl{The main motivation of this work is twofold: to increase the visual fidelity of driving simulations and reduce the manual labor requirements for 3D mesh and texture creation. With the use of GAN-based photorealistic image synthesizers, background objects such as trees, mountains, and the sky can be generated without detailed meshes or texture information.  However, conventional rendering is still needed to have full control over important objects of interest, such as vehicles and road markers.  }

In this paper, we propose to integrate generative adversarial image synthesis into a driving simulation. For each time step, CARLA, an open-source driving simulator \cite{dosovitskiy2017carla}, determines the scene's semantic layout with simple, textureless 3D models that are radiant with a unique class color. It should be noted that there is no illumination source other than the radiant 3D objects and no reflections or ambient occlusion are considered at this step. Then, a virtual pinhole camera is used to form a 2D semantic image from this scene. This image is the equivalent of a pixel-wise semantic segmentation mask. Next, the GAN-based image synthesizer converts the 2D semantic image to a photorealistic image. Conditional GAN (cGAN) \cite{mirza2014conditional} and CYcle GAN (Cy-GAN) \cite{isola2017image} are the main \hl{techniques} for this step. Simultaneously, a few objects of interest are partially rendered using a conventional rendering engine \cite{karis2013real}. This is necessary as full control over some critical objects, such as lane markings and vehicles in a driving scene, \hl{is} only achieved with a conventional graphics pipe.  Finally, a blending GAN mixes the cGAN/Cy-GAN synthesized image with the individually rendered objects. The proposed method was evaluated with \hl{semantic segmentation} \cite{chen2017deeplab}, an important driving-related perception task. 

\vspace{2mm}

\begin{figure}[t!]
	\centering
	\includegraphics[width=0.9\linewidth]{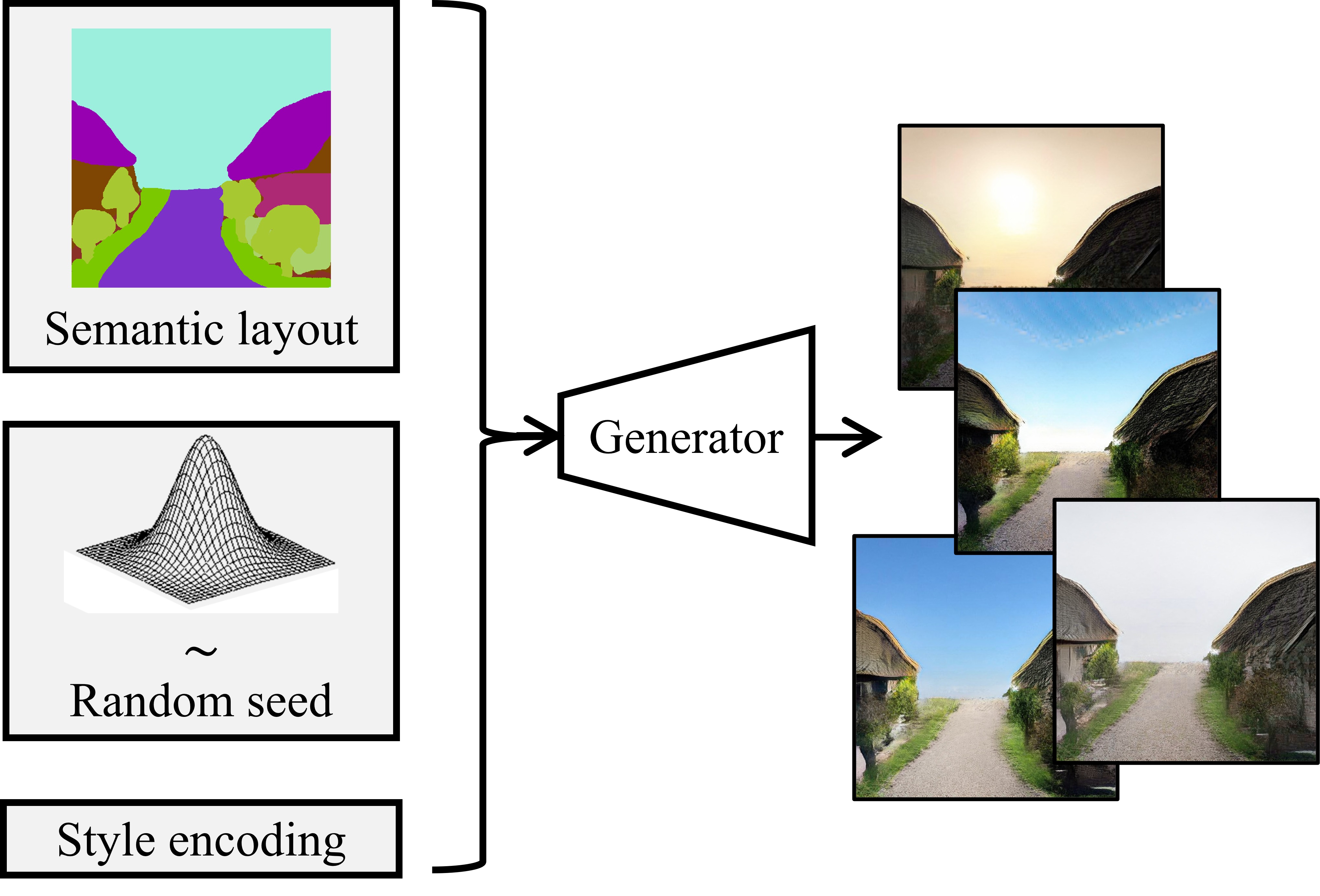}
	\caption{We first create the semantic layout, and then use SPADE \cite{park2019semantic} with different style encodings to generate random but photorealistic RGB \hl{background} imagery. Repetitive patterns that are common in driving simulations are memorazible by learning algorithms and break immersion for human driver subjects. The proposed approach alleviates these shortcomings. }
	\label{fig:gan_random}
\end{figure}

The main contributions of this work are:

\begin{itemize}
	\item  We introduce a novel driving simulation graphics pipeline for expediting scene creation using automated synthesis of background elements such as buildings, vegetation, and sky. To the best of our knowledge, this is the first GAN-render hybrid graphics pipeline for driving simulations.
	\item  Blending GAN-based image synthesis with physics-based partial rendering.
	\item Replacing \hl{recurring} patterns, \hl{such as repeating tree and building models,} that are common in driving simulations with generative photorealistic surfaces as shown in Figure \ref{fig:gan_random}. \hl{Repetitive patterns can break immersion for human-in-the-loop simulation experiments. In addition, machine learning algorithms trained or fine-tuned in a repetitive environment can fail in the real world due to overfitting.} As such, the proposed approach \hl{aims at} increasing the integrity of simulation-based\hl{ }intelligent transportation research.
	
\end{itemize}

\section{Related work}

\textbf{Simulation based driving studies.} Human driver reaction to various driving-related stimuli has been observed via simulation environments in numerous studies. The simulation's visual fidelity is critical for such experiments, as humans are accustomed to a real-world driving setting. \hl{Driving simulators have been used to study the driver's reaction during an automated driving take-over {\cite{minhas2020effects}}, to monitor human responses to stressful driving stimuli {\cite{lanata2014autonomic}}, to find the effect of inter-vehicular distances on human car following behavior {\cite{9295379}}, and to measure the effect of acoustic cues on situational awareness of human drivers {\cite{ju2020acoustic}}.}
\hl{An automated highway driving system with human-like decision-making capabilities} has been developed via a driving simulator \cite{yang2019automated}. Another study \cite{ludl2020enhancing} focused on human pose estimation using simulated images and showed that \hl{data-centric algorithms fine-tuned in these simulations} could be used in real-world scenarios.

\hl{A recent study showed that human subjects gaze with higher variance and exhibit more diverse steering activity in driving simulations that have better visual fidelity} \cite{van2015effects}. \hl{Higher visual fidelity is always desired in human-in-the-loop experiments because human driving behavior deviates from real-world behavior in unrealistic simulation environments} \cite{zhao2018use}. \hl{Furthermore, data-centric methods that are trained on synthetic data generated by conventional rendering engines fail to perform with real-world images} \cite{hoffman2018cycada}. 

The number and \hl{significance} of these studies underline the importance \hl{for} improving the visual fidelity \hl{in} driving simulations. \hl{New technologies can be developed more effectively with a better simulator. For example, if photorealism can be achieved, a learning-based lane-boundary detection algorithm} \cite{fan2019spinnet} \hl{can be trained in a simulation and deployed in the real world.}  

\textbf{Rendering.} Physics-based rendering \cite{kajiya1986rendering} has been used at the end \hl{of the line} of conventional computer graphics pipelines to form 2D imagery from virtual 3D scenes for a long time. The most common approaches, rasterization and ray-tracing, require a full pipeline of detailed 3D models, their surface textures and materials, and a physics engine such as Unreal Engine 4 \cite{karis2013real} to run complicated calculations for representing light and shading. Here, we propose to partially replace this pipe with much simpler 3D models and remove\hl{ }light, texture, and material information for most of the objects in the scene. We also show that \hl{the} visual fidelity can be increased with the proposed method.   


%

\textbf{Neural rendering.} \hl{Recent work }\cite{nguyen2018rendernet} demonstrated that 2D image formation could be achieved given a camera pose and light position in a 3D scene using differentiable convolutional networks. The key enabler here is the \hl{formulation of the discrete rasterization problem }\cite{liu2019soft}. With a differentiable rendering framework, a neural network can be trained with backpropagation. There is additional work \cite{gkioulekas2016evaluation, kato2018neural, loper2014opendr} focusing on the different aspects of differentiable rendering formulations and approximations. Neural rendering \hl{is a promising technique.} However, this approach still requires detailed 3D models and is incapable of generating texture information, which reduces the visual fidelity of the output. \hl{In comparison, we propose to use generative models for reducing 3D model and texture complexity.}

\textbf{Generative adversarial image synthesis.} \hl{Generative adversarial image synthesis omits rasterization and rendering.} Physical phenomena such as lighting and reflectivity are completely ignored by  GAN based neural image synthesizers \cite{brock2018large,zhang2018self,zhang2017stackgan,zhang2018stackgan++,bau2018gan,liu2017unsupervised,miyato2018cgans}. Instead, the photorealism is achieved by training the GAN with real-world data. In other words, the network learns to generate photorealistic images \hl{by capturing a latent probability distribution underlying real-world datasets.} This approach has one major drawback: there is no constraint on the semantic layout of the generated 2D image. Hence, no association with 3D scenery can be constructed.  As such, this methodology cannot be applied for our image formation purposes.   

\textbf{Conditional generative adversarial image synthesis.} On the other hand, conditional GANs \cite{isola2017image,chen2017photographic,park2019semantic,qi2018semi,wang2018video, wang2018high}, \cite{wang2019example} have been effectively used for image synthesis while retaining a semantic constraint. Typically, this constraint is a pixel-wise semantic segmentation mask, but other modalities such as text \cite{reed2016generative} have also been used. One limiting factor for \hl{Conditional GAN (cGAN)} is the paired data requirement. The dataset must contain semantic segmentation masks and the corresponding real-world images. Building such paired datasets is labor-intensive \hl{because every real-world image needs a corresponding semantic segmentation label assigned by a human annotator.}


\textbf{Cycle-consistency and domain adaptation.} Cycle consistent GANs and unsupervised domain adaptation techniques \hl{make the paired dataset requirement uneccessary} \cite{deng2018image,hoffman2018cycada, bousmalis2017unsupervised, liu2016coupled, zhu2017unpaired, tzeng2017adversarial, bousmalis2017unsupervised}. These works have illustrated that high fidelity \hl{image synthesis can also be achieved with unpaired data. Cycle-consistency} is very promising and has a huge application range. For example, CyCADA \cite{hoffman2018cycada} can translate  an existing game-engine generated image into a photorealistic image\hl{.} 




\hl{T}he aforementioned GAN-based image synthesis techniques have not been integrated into driving simulation pipelines until now. \hl{This contribution makes our proposed method novel.} We propose to use simple 3D models radiant with unique class color-codes without textures to form a 2D semantic image\hl{.} This image is analogous to a 2D semantic segmentation mask. Then, a state-of-the-art GAN-based image \hl{synthesizer} trained on real-world datasets \hl{is used} to generate RGB imagery. We tried both cGAN and Cy-GAN variants. Additionally, we render certain important objects of interest, such as cars in an urban scene, with Unreal Engine 4. \hl{Images obtained by blending the partial-render foreground and GAN background are more realistic. Blended images also retain the semantic layout of the scene better.}
%

\section{Preliminaries}

Generative Adversarial Networks (GAN) \cite{goodfellow2014generative} use a generator $G$ and a discriminator $D$ in a simultaneous adversarial training strategy. The \hl{goal} of $G$ is \hl{to generate} data $\hat{x}$ that is indistinguishable from the real data $x \in X$. During training, $G$ captures the \hl{probability distribution $p_{\textrm{data}}$ which should closely match the distribution underlying the real data.} This is achieved by training a generative mapping function $G(z)$ that maps\hl{ }an a priori noise distribution $p_{z}(z)$ to the data domain $X$. While $G$ tries to generate the most realistic $\hat{x}$, the discriminator $D$ tries to discriminate fake data $\hat{x}$ from real data $x$. The output of $D(x)$\hl{ }is the probability that $x$ is real.
$G(z)$ and $D(x)$, both of which are neural networks, \hl{are trained} simultaneously with the following min max \hl{function}:  

\begin{align}
\label{eq:minmaxGAN}
\begin{split}
\underset{G}{\textrm{min}}\underset{D}{\textrm{max}}V(D,G) & = \mathbb{E}_{\boldsymbol{x} \sim p_{\textrm{data}}(\boldsymbol{x})} \left[ \textrm{log}D(\boldsymbol{x}) \right] + \\
& \mathbb{E}_{\boldsymbol{z} \sim p_{z}(\boldsymbol{z})} \left[ \textrm{log}(1-D(G(\boldsymbol{z}))) \right] .
\end{split}
\end{align}

\section{Method}


\begin{figure*}[t!]
	\centering
	\includegraphics[width=1\linewidth]{fig/overview1}
	\caption{Overview of the proposed method. We introduce a novel neural graphics pipeline to form 2D image representations from virtual 3D scenes. Most of the scene is generated with very simple 3D models without texture except for a few partially rendered objects of interest. We then blend the cGAN synthesized image with a physics-based partial render for increasing visual fidelity \textit{and} to maintain full control over the appearance of objects of interest.}
	\label{fig:overview}
\end{figure*}

\subsection{Problem formulation}

We define a virtual 3D driving scene $S$ with a 6-tuple $(O_{1}, O_{2}, P_{1}, P_{2}, T_{2}, \textbf{x})$. Where $O_{1} = \left( \textbf{o}_{1, 1}, \textbf{o}_{1, 2}, \cdots, \textbf{o}_{1, n} \right) $ is a list of object pose vectors, $\textbf{o} \in \mathbb{R}^{6}$, and  $P_{1} = \left( M_{1}, M_{2}, \cdots, M_{n} \right) $ is the list of corresponding simple object meshes. We assume $P_{1}$ is radiant with unique class color-codes. $O_{2}$ is a sublist of $O_{1}$ for certain objects of interest, and it has a corresponding list of more complicated object meshes $P_{2}$. $P_{2}$ is not radiant. $T_{2}$ is a list of texture maps that corresponds to $P_{2}$. $\textbf{x} \in \mathbb{R}^{6}$ is the pose vector of a virtual camera. It should be noted that a corresponding $T_{1}$ to $O_{1}$ does not exist.

We follow the formal definition of a triangular mesh given in \cite{botsch2010polygon}. $M := (V,Q)$ is a triangular mesh defined with faces $Q \subseteq \left\lbrace 1, \cdots, |V|\right\rbrace  ^{3}$ and vertices $V \subseteq \mathbb{R}^{3}$, where $q = (q_{1}, q_{2}, q_{3}) \in Q$ is a triangular face with corresponding vertices $v_{q_{1}}$, $v_{q_{2}}$, and $v_{q_{3}}$. $E(Q)$, the edges between the vertices, are defined by the faces implicitly.


\textbf{Problem 1.} Given $S$, we are interested in finding a mapping function 
$U: \textbf{x} \rightarrow \mathbb{R}^{H\times W \times 3}$ that will convert  the camera pose vector $\textbf{x}$ to a photo-realistic RGB image with height $H$ and width $W$.

The overview of our solution, Hybrid Generative Neural Graphics (HGNG) is shown in Figure \ref{fig:overview} and Algorithm 1, and the formal description follows.

\subsection{Semantic Image formation}

\begin{algorithm}[htp]
	\SetAlgoLined
	\KwIn{\\$O_{1}$, the list of object pose vectors\\
		$P_{1}$, the list of simple object meshes w/o texture. \\
		$O_{2}$, a sublist of $O_{1}$, corresponds to objects of interest \\	
		$P_{2}$, the list of complex object meshes. \\
		$T_{2}$, the list of texture maps that corresponds to $P_{2}, O_2$. \\
		$\textbf{x}$, the pose vector of the pinhole camera
	}
	
	\KwOut{\\ $ \textbf{I} \in \mathbb{R}^{H\times W \times 3}$, 2D RGB image.}
	{\bf Main algorithm:} \\
	$\textbf{m}=h_\text{pinhole}(O_{1},P_{1},\textbf{x})$\;
	$\textbf{I}_\text{background}=f_\text{generator}(\textbf{m},z \sim N)$\;
	
	\ForEach{$i(1,2...n)$}{
		$\textbf{I}_{i}^\text{object-of-interest} = L_\text{rendering}(O_{2}(i), P_2(i), T_2(i))$\;
	}
	$\textbf{I}_\text{foreground}=\sum_{i}^{n}\textbf{I}_{i}^\text{object-of-interest }$\;
	$\textbf{I}=b_\text{generator-blending}(\textbf{I}_\text{background}, \textbf{I}_\text{foreground})$\;
	\caption{HGNG($O_{1}, O_{2}, P_{1}, P_{2}, T_{2}, \textbf{x}$)}
	\label{algo}
	
\end{algorithm}

A semantic image formation function $h$ can be obtained with $O_{1}$, $P_{1}$ and a pinhole camera model. 
\hl{Let $\textbf{m} \in \mathbb{M} ^{H \times W}$ be a pixel-wise semantic image whose entries correspond to the semantic classes of the scene. Then}  
$h: \textbf{x} \rightarrow \mathbb{M} ^{H \times W}$ maps $\textbf{x}$ to an integer subspace ($\mathbb{M} \subset \mathbb{Z}$) using the pinhole camera model \cite{hartley2003multiple} given by:

\begin{align}
\label{eq:pinhole}
\begin{pmatrix}
m_{1} \\
m_{2}
\end{pmatrix}
= -\dfrac{d}{p_{3}}\begin{pmatrix}
p_{1} \\
p_{2}
\end{pmatrix}
\end{align}
%
%
where $(p_{1}, p_{2}, p_{3})$ \hl{are} the 3D coordinates of point $\textbf{p}$ in $\mathbb{R}^{3}$, $(m_{1},m_{2})$ \hl{are} the corresponding pixel coordinates in $\textbf{m}$, and $d$ is the distance between the focal point and image formation plane.  $\textbf{m}$ is an upside-down image as shown in Figure \ref{fig:overview}. \textbf{m} is rotated $180^{\circ}$ for the next step. For simplicity, we use the same notation $\textbf{m}$ for the rotated image in the remainder of the paper.

Then, the problem narrows down to finding $f: \textbf{m} \rightarrow \mathbb{R}^{H \times W \times 3}$. This is the exact same goal as the well-studied \cite{isola2017image,chen2017photographic,park2019semantic,qi2018semi} conditional image synthesis problem.

\subsection{Generative Adversarial Image Synthesis with cGANs and Cy-GANs}

We propose to use the generator networks of cGANs or Cy-GANs to map $G: \textbf{m} \rightarrow \mathbb{R}^{H \times W \times 3}$. Training is to be done on a real-world paired dataset of  RGB images and pixel-wise semantic masks for cGAN. On the other hand, Cy-GANs can be trained with an unpaired dataset.  

cGAN \cite{mirza2014conditional} extends the original GAN and can generate realistic fake data while retaining a conditional constraint. This is achieved by pairing the conditional constraint $y$ with the data $x$ and creating a new paired dataset $(x,y)$. This pair can be an RGB image and pixel-wise semantic layout pair, or an image and text pair. $x$ and $y$ do not have to share the \hl{same} modality. Details of cGAN can be found in \cite{mirza2014conditional}. cGAN can successfully generate photo-realistic fake data with a conditional constraint. However, the paired dataset requirement increases the cost of this approach.

\begin{figure*}[t!]
	\centering
	\includegraphics[width=1\linewidth]{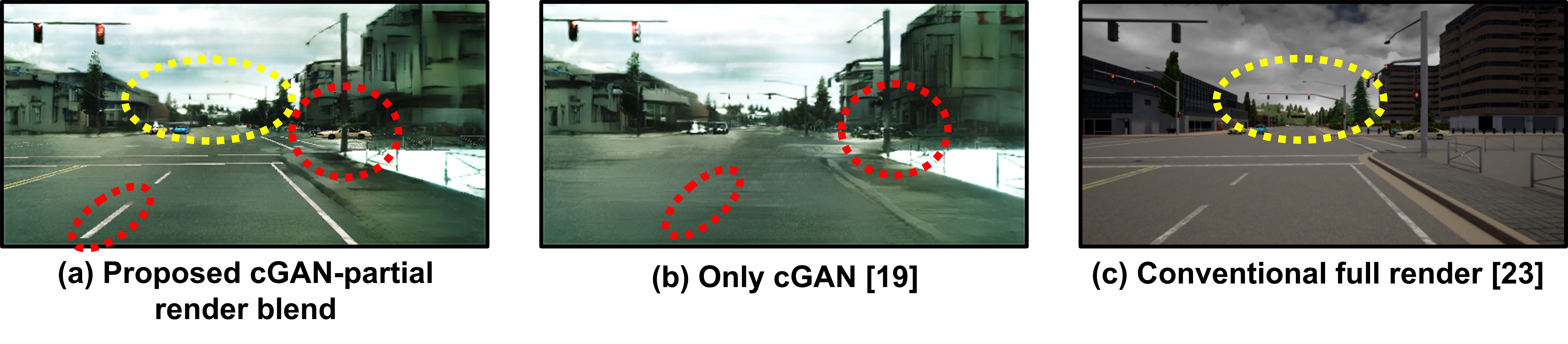}
	\caption{\hl{The proposed framework (a) converts the semantic layout of the scene into a photorealistic image by blending partially rendered foreground objects with a GAN generated background. The conventional rendering engine}  \cite{dosovitskiy2017carla} \hl{(c) requires detailed models and texture information while outputting unrealistic background trees and vegetation (shown with a yellow circle). On the other hand, using only a cGAN (b)}  \cite{park2019semantic} \hl{ approach leads to poor car shapes and omitting road markings (shown with a red circle), while removing the need for texturing and rendering calculations. The proposed method (a) has the best of both worlds.}  }
	\label{fig:comparison}
\end{figure*}



In comparison, building an unpaired $X$ and $Y$ is relatively easy. Cycle GAN (Cy-GAN) \cite{zhu2017unpaired} enables photo-realistic image synthesis with unpaired data. In summary, Cy-GAN contains two generators, $G(x)$ and $F(y)$, which map $X \rightarrow Y$ and $Y \rightarrow X$ respectively. Also, two discriminators, $D_{X}$ and $D_{Y}$, try to distinguish fake data from real data. The adversarial losses are similar to the original GAN; the addition is the novel cycle consistency loss. This loss prevents the mappings of $G$ and $F$ from diverging from each other. The key idea of cycle GAN is \hl{the use of }two generators to create a cycle. First, $G(x)$ generates fake $\hat{y}$, then $F(G(x))$ translates the fake  $\hat{y}$ back to $\hat{x}$. If the cycle is consistent, then $x \approx \hat{x}$. 


The baseline cGAN employed in this study is a SPatially-Adaptive-(DE)-normalization (SPADE) \cite{park2019semantic} network, which is a state-of-the-art cGAN based image synthesizer. SPADE outperforms other image-to-image synthesizers by retaining semantic information against conventional normalization operations \cite{park2019semantic}. This is achieved through the following de-normalization operation where the activation value at layer $i$ is given by:

\begin{align}
\gamma_{c,y,x}^{i}(\textbf{m}) \dfrac{h^{i}_{n,c,y,x} - \mu_{c}^{i}}{\sigma^{i}_{c}}  + \beta^{i}_{c,y,x}(\textbf{m})
\end{align}
\noindent where $h^{i}_{n,c,y,x}$ is the activation before normalization and  $\mu_{c}^{i}$ and $\sigma^{i}_{c}$ are the mean and standard deviation in channel $c$. $\gamma_{c,y,x}^{i}(\textbf{m})$ and $ \beta^{i}_{c,y,x}(\textbf{m})$ are learned \hl{variables} that modulates the normalization process. We refer the readers \hl{of} the original SPADE paper \cite{park2019semantic} for more details.

We use a SPADE network pre-trained on the Cityscapes dataset \cite{cordts2016cityscapes} as the mapping function $f_{s}$ and obtain the synthesized image with it as $\textbf{I} = f_{s}(\textbf{m}) $.

\subsection{Partial rendering}

To increase visual fidelity and have full control over certain objects of interest, we propose using physics-based rendering to obtain partially-rendered images $\textbf{I}_{r}$. Besides $O_{2}, P_{2}, T_{2}$ and $\textbf{x}$, a light source is also needed for rendering. Here we assume \hl{that} the properties and location of the light source are fixed and known relative to $\textbf{x}$. Then the rendering equation \cite{kajiya1986rendering} can be used to render objects of interest.  

\begin{align}
\label{eq:rendering}
\begin{split}
L_{0}(\textbf{p},\omega, \lambda, t ) & = L_{e}(\textbf{p},\omega_{0}, \lambda, t) + \\
& \int_\Omega f_{r}(\textbf{p},\omega_{i},\omega_{0} \lambda, t)L_{i}(\textbf{p},\omega_{i}, \lambda, t)(\omega_{i}.\textbf{n})d\omega_{i}
\end{split}
\end{align}
where $L_{0}(\textbf{p},\omega, \lambda, t )$ is the total spectral radiance, $\lambda$ is wavelength, $\omega_{0}$ is the outgoing light direction, $\omega_{i}$ is the incoming light direction, $t$ is time and $\textbf{p} $ is a point in 3D space. $L_{e}(\textbf{p},\omega_{0}, \lambda, t)$ is the emitted spectral radiance, $\Omega$ is a unit hemisphere with the surface normal center $\textbf{n}$ of $\textbf{p}$ and it contains all values for $\omega_{i}$,  $f_{r}(\textbf{p},\omega_{i},\omega_{0} \lambda, t)$ is the bidirectional reflectance function and finally $L_{i}(\textbf{p},\omega_{i}, \lambda, t)$ is the spectral radiance of the incoming wavelength.   

With Equation \ref{eq:rendering}, the spectral radiance of each 3D point on a few objects of interest is obtained. Then, the partially rendered image $\textbf{I}_{r}$ is formed with the same pinhole camera model introduced in Equation \ref{eq:pinhole}.

\subsection{Blending}



Here we propose to blend the synthesized image $\textbf{I}$ with the partially rendered image $\textbf{I}_{r}$ to obtain a hybrid image $\textbf{I}_{h}$ as shown in Figure \ref{fig:overview}. The hybrid image is defined as:

\begin{figure*}
	\centering
	\includegraphics[width=0.9\linewidth]{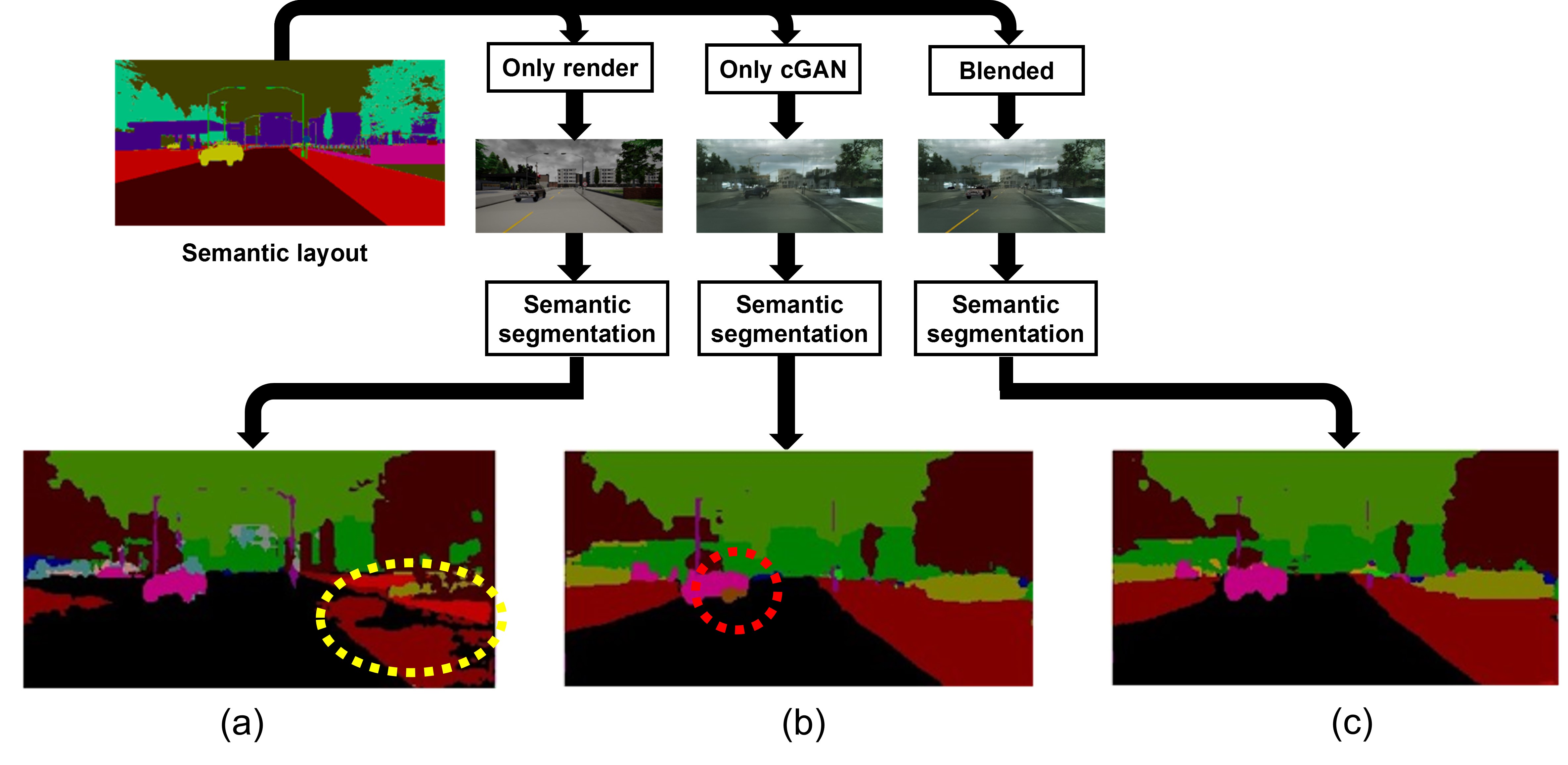}
	\caption{An illustration of semantic retention analysis. The semantic segmentation result should stay true to the initial semantic layout.  (a) Full-render yields unrealistic shadows. On the bottom right-hand side of the left-most image \hl{(shown with a yellow circle)},  shadows of trees cast on the sidewalk were misclassified as a road by DeepLabV3. (b) cGAN generated vehicles do not retain their shapes perfectly \hl{(middle image, shown with a red circle)}. (c) Blending retains the semantic relationship with the source layout (right-most image). This figure employs different color codes to distinguish the semantic layout formation and semantic segmentation processes for illustration purposes.}
	\label{fig:retention}
\end{figure*}

\begin{align}
\textbf{I}_{h} := b(\textbf{I},\textbf{I}_{r}) 
\end{align}
where the blending function $b: (\textbf{I},\textbf{I}_{r}) \rightarrow \mathbb{R}^{HxW \times 3}$ maps the synthesized and partially rendered images to a new hybrid RGB image. We compared three different blending functions $b$ in this study.

\textbf{Alpha blending.} Taking $\textbf{I}$ as the background image and $\textbf{I}_{r}$ as the foreground image, the alpha blended image $\textbf{I}_{h}$ can be obtained with:

\begin{align}
\label{eq:alpha_blend1}
\textbf{I}_{h} = \alpha\textbf{I} +(1-\alpha)\textbf{I}_{r}.
\end{align}

\textbf{Pyramid blending.} With the gaussian pyramid mask $G_{R}$ \cite{mertens2009exposure}, $L_{a}$ the laplacian pyramid of the foreground $\textbf{I}_{r}$, and $L_{b}$ the laplacian pyramid of background $\textbf{I}$, the laplacian blended pixel $b(i,j)$ can be obtained with:  

\begin{align}
\label{eq:pyramid_blend1}
b(i,j) = G_{R}(i,j)L_{a}(i,j) + (1-G_{R}(i,j))L_{b}(i,j).
\end{align}

\textbf{GAN blending.} As a third blending option, we employed GP-GAN \cite{wu2019gp}. The generator of GP-GAN converts a naive copy-paste blended image to a realistic well-blended image. Besides conditional GAN loss, GP-GAN employs an auxiliary $l_{2}$ loss to sharpen the image.
The overall combined loss function is given by:
\begin{align}
\label{eq:gpgan1}
\mathcal{L}(x, x_{g}) = \lambda \mathcal{L}_{l_{2}}(x,x_{g}) + (1-\lambda) \mathcal{L}_{\textrm{adv}}(x,x_{g})
\end{align} 
where $\mathcal{L}(x, x_{g})$ is the final loss, $\mathcal{L}_{l_{2}}$ is the $l_{2}$ loss and $\mathcal{L}_{\textrm{adv}}$ is the adverserial loss. $\lambda$ is a hyperparameter and set to 0.999. 



%
%

\section{Experiments}\label{sec:exp}

\subsection{Implementation details}

We used SPADE \cite{park2019semantic} as our cGAN image synthesizer to convert the semantic layout of the scene to a photorealistic background image. The network was trained on Cityscapes \cite{cordts2016cityscapes}, an urban driving dataset with paired semantic mask and image data. CARLA \cite{dosovitskiy2017carla}, an open-source driving simulator built upon Unreal Engine 4, was utilized to obtain the semantic layout and partially rendered images. We used the shading and lighting engine \cite{karis2013real} of Unreal Engine 4 in our experiments. Only vehicles and lane markings were considered as objects of interest. For blending, we used a GP-GAN \cite{wu2019gp} trained on the Transient Attributes Database \cite{laffont2014transient}. All computational experiments were conducted with an Nvidia RTX 2080.



\begin{figure*}[t!]
	\centering
	\includegraphics[width=1\linewidth]{fig/corr.jpg}
	\caption{InceptionV3 feature vector correlation matrices of real and synthetic data. \hl{The synthetic dataset that was generated with the proposed blending approach shows a similar  correlation pattern with real data. This pattern does not emerge with the only render or only GAN methods. }} 
	\label{fig:corr}
\end{figure*}
\subsection{Evaluation}


\subsubsection{Semantic retention} Figure \ref{fig:retention} illustrates semantic retention analysis, a common \cite{chen2017photographic,wang2018high,park2019semantic} evaluation method for fake image synthesis. Semantic retention measures the semantic correspondence between the conditional semantic mask and the synthesized image. In summary, an external semantic segmentation network is used to segment the synthesized image. Then, the discrepancy between the conditional semantic layout (input of the synthesizer) and the semantic mask obtained from the generated image (output of the pre-trained external segmentation network) is calculated with top-1 accuracy. A good synthesizer should produce photorealistic images while retaining the initial conditional semantic layout. In other words, the initial semantic layout is accepted as the ground truth, and the image synthesizer's mask accuracy is calculated to obtain the retention score. A higher retention score is favorable.

In this study, we employed DeepLabV3 \cite{chen2017deeplab}, a state-of-the-art semantic segmentation network, to measure semantic retention. The network  was trained on Cityscapes, an urban driving dataset \cite{cordts2016cityscapes}. 

\subsubsection{FID} Frechet Inception Distance (FID) \cite{heusel2017gans} is a commonly used \cite{park2019semantic, wang2018video} performance metric for measuring visual fidelity. In summary, a deep neural network is employed to extract features of all images in a dataset. Then, the covariance and mean of features obtained from synthesized and real datasets are compared to generate a score. We do not have any real-data corresponding to our virtual 3D scene, but FID can still be used with unpaired data. As such, three different real-world datasets \cite{cordts2016cityscapes,geiger2013vision, zhou2017scene} were utilized as the ground truth. 

An InceptionV3 \cite{szegedy2016rethinking} model that was trained on ImageNet \cite{deng2009imagenet} was employed as the feature extractor. After features were extracted from the synthesized images and from real-world images from Cityscapes \cite{cordts2016cityscapes}, KITTI \cite{geiger2013vision}, and ADE20K \cite{zhou2017scene}, the FID is calculated as follows:

\begin{align}
\label{eq:fid}
d^2 = ||\mu_1 - \mu_2||^2 + Tr(C_1 + C_2 - 2\sqrt{(C_1C_2)})
\end{align} 
where $\mu_{1}, \mu_{2}$ are the means of features, and $C_{1},C_{2}$ are the covariances obtained from datasets $1$ and $2$ respectively, where the first dataset consists of real images and the second synthesized images. The smaller the distance $d^{2}$, the more similar are the two datasets. In other words, a small FID indicates that fake data is similar to real-world data.

The synthesized images were then compared against each other using FID scores as shown in Table \ref{table:FID}. $\mu_{1}$ and $C_{1}$ were obtained from the real datasets and do not change in a column, whereas $\mu_{2}$ and $C_{2}$ were obtained from synthesized images and vary with each row.  A lower FID indicates high visual fidelity.



\subsubsection{Inception score} Inception Score (IS) was initially proposed to evaluate the generator performance of GANs  \cite{salimans2016improved}. In summary, a pre-trained image classifier is run over a GAN generated fake dataset. The distribution of predicted classes, along with the confidence intervals, were then compared against a real dataset. A higher Inception Score indicates higher image quality and diversity. IS differs from FID by its use of actual classification results, whereas FID utilizes latent features. Details of IS can be found in \cite{salimans2016improved}.

\subsubsection{Comparisions and ablations} Ablation studies were conducted to demonstrate the effect of each component of the proposed method. The ablation list is:


\begin{enumerate}
	\item No partial-render (only vanilla cGAN or Cy-GAN)
	\item No cGAN or Cy-GAN (only full render)
	\item Alpha blend (cGAN or Cy-GAN + partial render)
	\item Pyramid blend (cGAN or Cy-GAN + partial render)
	\item GAN blend (cGAN or Cy-GAN + partial render)
\end{enumerate}

Where SPADE \cite{park2019semantic} was used as the vanilla cGAN and the original Cycle-GAN \cite{isola2017image} was employed as the vanilla Cy-GAN variant.  


We used CARLA \cite{dosovitskiy2017carla} to obtain fully rendered images of urban scenery. The semantic layout of the scene was also imported from CARLA and used as the conditional input for the generative adversarial image synthesizers. Only vehicles and lane markings were considered by the partial-renders. In this work, the image synthesis was done frame-by-frame with a fixed random seed. 



\subsection{Results}

\subsubsection{Qualitative results}

The qualitative results are shown in Figures \ref{fig:comparison}, \ref{fig:retention}, and \ref{fig:corr}. These figures illustrate fully rendered, blended, and only cGAN images. As can be seen in Figure \ref{fig:retention}, rendered shadows are unrealistic, while only cGAN generated vehicles cannot retain their shapes. These results underline the importance of partial rendering of objects of interest such as cars, vans, and lane markings. The hybrid approach combines the accuracy of a full-render with the realism of a generative model.

\hl{Treating foreground objects differently from background scenery with the proposed blending technique improves the photorealism of the final image as can be seen in Figures} \ref{fig:comparison}, \ref{fig:retention}, \ref{fig:corr}. 
\hl{The appearance of foreground objects, such as vehicles, needs to be controllable and rendered in detail. This necessitates employing conventional rendering techniques with high-detail models and textures. However, background sceneries cannot be controlled at the same level of detail because they contain a higher number of elements such as mountains, trees, and buildings. In practice, conventional driving simulators pay less attention to these background elements by lowering 3D model quality and using less rendering focus. This causes lower overall visual fidelity. In contrast, GAN-based image synthesizers can automate background scene generation while achieving higher visual fidelity by learning the background compositions of real-world data. Our use of a GAN-based image synthesizer completely removes the texture and detailed model requirements of background scenery generation while increasing visual fidelity. In Figure} \ref{fig:comparison}, \hl{the conventional rendering method produced unrealistic trees and vegetation (shown with a yellow circle) around the vanishing point of the image. At the same time, the proposed method and the only-cGAN approach generated a more blended background scene with vegetation at the same spot.}

\begin{table}
	\begin{center}
		\caption{Semantic retention performance- higher scores are better. Our methods outperform the physics-based rendering approach.}
		\label{table:retention1}
		\begin{tabular}{|l|c|}
			\hline
			\hspace{1cm} Method  &  Semantic retention $\uparrow$ \\
			\hline
			\textbf{Baseline:} only render \cite{dosovitskiy2017carla}    & 0.819 \\
			only Cycle GAN \cite{isola2017image}  & 0.343 \\
			\hline
			\textbf{Proposed} & \\
			\indent $ $ cy-GAN alpha blend & 0.362  \\
			\indent $ $ cy-GAN pyramid blend & 0.353 \\
			\indent $ $ cy-GAN GAN blend & 0.318   \\
			\indent $ $ cGAN alpha blend & 0.879   \\
			\indent $ $ cGAN pyramid blend  & 0.868   \\ 
			\indent $ $ cGAN GAN blend  & 0.846   \\              
			
			\hline
		\end{tabular}
	\end{center}
\end{table}

\begin{table}
	\begin{center}
		\caption{FID performance- lower scores are better. Our methods outperform the physics-based computer graphics pipeline. Cy-R stands for CyGAN-Render blend, and c-R stands for cGAN-Render blend.}
		\label{table:FID}
		\begin{tabular}{|l|c|c|c|}
			\hline
			& \multicolumn{3}{c|}{FID $\downarrow$} \\
			\cline{2-4} 
			\hspace{0.7cm} Method &  Cityscapes\cite{cordts2016cityscapes}& KITTI\cite{geiger2013vision} & ADE20K\cite{zhou2017scene} \\
			\hline
			Only render \cite{dosovitskiy2017carla} & 231.768 & 285.222 &  361.496\\
			\hline
			\textbf{Proposed} & & &\\
			\indent $ $ Cy-R alpha blend & 175.832 & 220.223 & 272.069\\
			\indent $ $ Cy-R pyramid blend & 196.911 & 228.277 &  279.704\\
			\indent $ $ Cy-R GAN blend & \textbf{194.191} & 234.087 & 266.615 \\
			\indent $ $ c-R alpha blend & 188.809 & 220.161 & 272.877 \\
			\indent $ $ c-R pyramid blend & 202.120 & \textbf{214.488} & 265.603\\ 
			\indent $ $ c-R GAN blend  & 194.898 & 217.663 & \textbf{260.404}\\                
			
			\hline
		\end{tabular}
	\end{center}
\end{table}

\hl{On the other hand, only using a GAN-based image synthesizer reduces control over the appearance of objects of interest, such as cars. This causes lower visual fidelity. In Figure} \ref{fig:comparison}, \hl{the only GAN-based approach failed to generate road-markings. In addition, the surface quality of cars (shown with a red circle) was much lower than the full-render approach and the proposed method. Figure} \ref{fig:retention} \hl{demonstrates the unrealistic shadows of full-render and incomplete vehicle shapes of the only GAN approach with a yellow and red circle. The proposed method alleviates these issues. These results indicate a qualitative validation of our hypothesis: the proposed approach, blending GAN-based synthesizers with conventional rendering, has the best of both worlds.}


\begin{figure}[t!]
	\centering
	\includegraphics[width=1\linewidth]{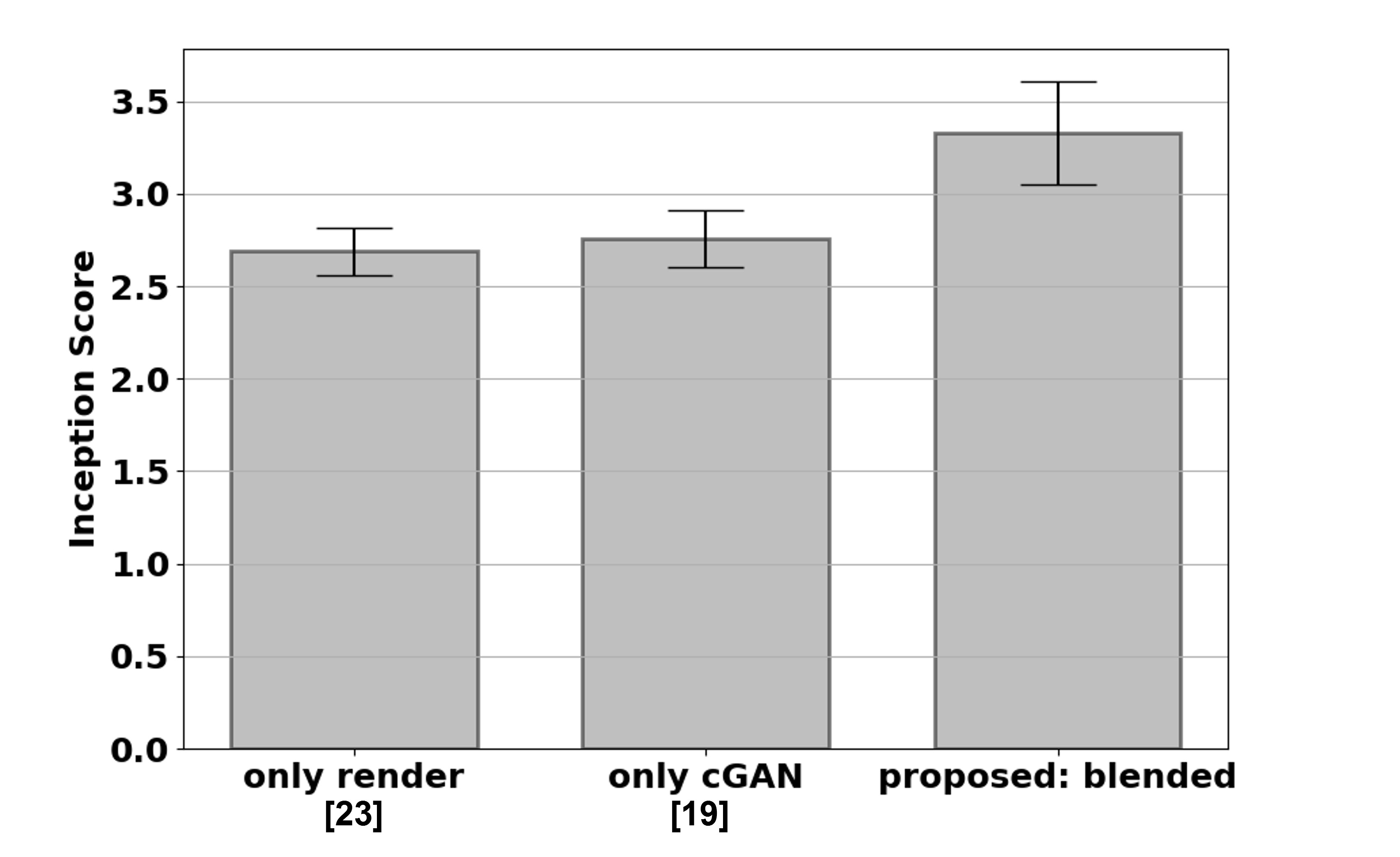}
	\caption{Inception score \cite{salimans2016improved}. A high Inception score indicates better image quality and higher diversity.  }
	\label{fig:inception_score}
\end{figure}

\subsubsection{Quantitative results}

Figure \ref{fig:corr} \hl{shows Inception V3 feature vector correlations. Even though real-world images of the Cityscapes and KITTI datasets are entirely different, they have similar latent feature correlations. However, this pattern does not emerge with a synthetic dataset of low visual fidelity. The proposed blended synthetic dataset has, albeit being weak, a similar correlation pattern. In comparison, the same pattern does not emerge with the conventional render or pure GAN approaches. This shows that the proposed blending approach is a good strategy for the realistic representation of driving scenes.}

IS, FID, and semantic retention scores are given in Table \ref{table:retention1}, Table \ref{table:FID} and Figure \ref{fig:inception_score}. These results indicate that the proposed hybrid blending approach consistently outperforms conventional rendering and pure generative adversarial image synthesis. Using detailed models with a conventional rendering engine for objects-of-interest produces high-quality visuals. However, building the rest of the driving scene with this level of detail is extremely challenging. Our hybrid method mitigates this problem by replacing the background elements with a GAN synthesizer and blending high-quality objects of interest renders into the scene. This blending strategy is the main reason for achieving higher visual fidelity.

The Cityscapes dataset contains only urban driving scenes, while ADE20K also has miscellaneous scenes. All of our virtual 3D scenes were in an urban environment. As such, most of the methods received better FID scores for the Cityscapes dataset, as can be seen in Table \ref{table:FID}. 

GAN blend and Alpha blend showed similar performances as shown in Table \ref{table:retention1} and Table \ref{table:FID}. However, it should be noted that the blending GAN was not trained on an urban driving dataset. The blending performance can possibly be increased with a better blending dataset for training the blending GAN. 

The cGAN variants performed better on average as expected, as shown in Table \ref{table:retention1} and \ref{table:FID}. The synthesized images were both realistic and loyal to the initial semantic layout. However, cGAN requires a paired dataset for training. The full render is better at semantic retention than Cy-GAN variants, but Cy-GAN variants have a higher FID score than rendering. This means that Cy-GAN can generate realistic images but fails to retain the semantic constraints.

\section{Conclusions}

This work introduced and investigated the feasibility of Hybrid Generative Neural Graphics (HGNG). \hl{The proposed approach utilizes a GAN-based image synthesizer to remove the need for rendering calculations and labor-intensive texture-making steps for background elements while increasing photorealism. In addition, our method achieves full control over the appearance of objects of interest using partial-rendering. Our novel image formation strategy blends the GAN-generated background image with these partial renders and outperforms conventional approaches. Experimental results }indicate that conventional generation of driving simulation graphics now has a strong alternative. 

\hl{In order to train the cGAN-based synthesizers, real-world urban images and their semantic labels, i.e., a paired dataset, are needed. Therefore, with the publication of more paired real-world datasets, the performance of the proposed method can be further increased. On the other hand, CyGANs remove this paired dataset requirement with the use of cycle consistency, but cyGANs do not perform as well as cGANs. As such, without a paired dataset, the proposed system cannot outperform the conventional pipelines yet.  However, potential future developments in domain adaptation and cycle consistency can greatly benefit HGNG and may remove the paired dataset requirement in the future.}

This work focused on frame-by-frame image formation with GANs. However, computer graphics applications such as driving simulations may require more temporally consistent approaches. \hl{Each subsequent frame of a driving simulation needs to be consistent with the overall sequence. The proposed method already achieves temporal consistency for objects of interest using partial rendering. The temporal consistency of the GAN-generated background scene can potentially be increased with larger urban video datasets. To this end, future work can focus on creating better urban video datasets and developing GAN-based video-to-video synthesis methods.  }

\section*{Acknowledgment}

Material reported here was supported by the United States Department of Transportation under Award Number 69A3551747111 for the Mobility21 University Transportation Center. Any opinions, findings, conclusions, or recommendations expressed herein are those of the authors and do not necessarily reflect the views of the United States Department of Transportation.

\ifCLASSOPTIONcaptionsoff
  \newpage
\fi



\bibliographystyle{IEEEtran}
\bibliography{AAAI_image_synthesis}
%

%

\vspace{-0.7cm}

\begin{IEEEbiography}[{\includegraphics[width=1in,height=1.25in,clip,keepaspectratio]{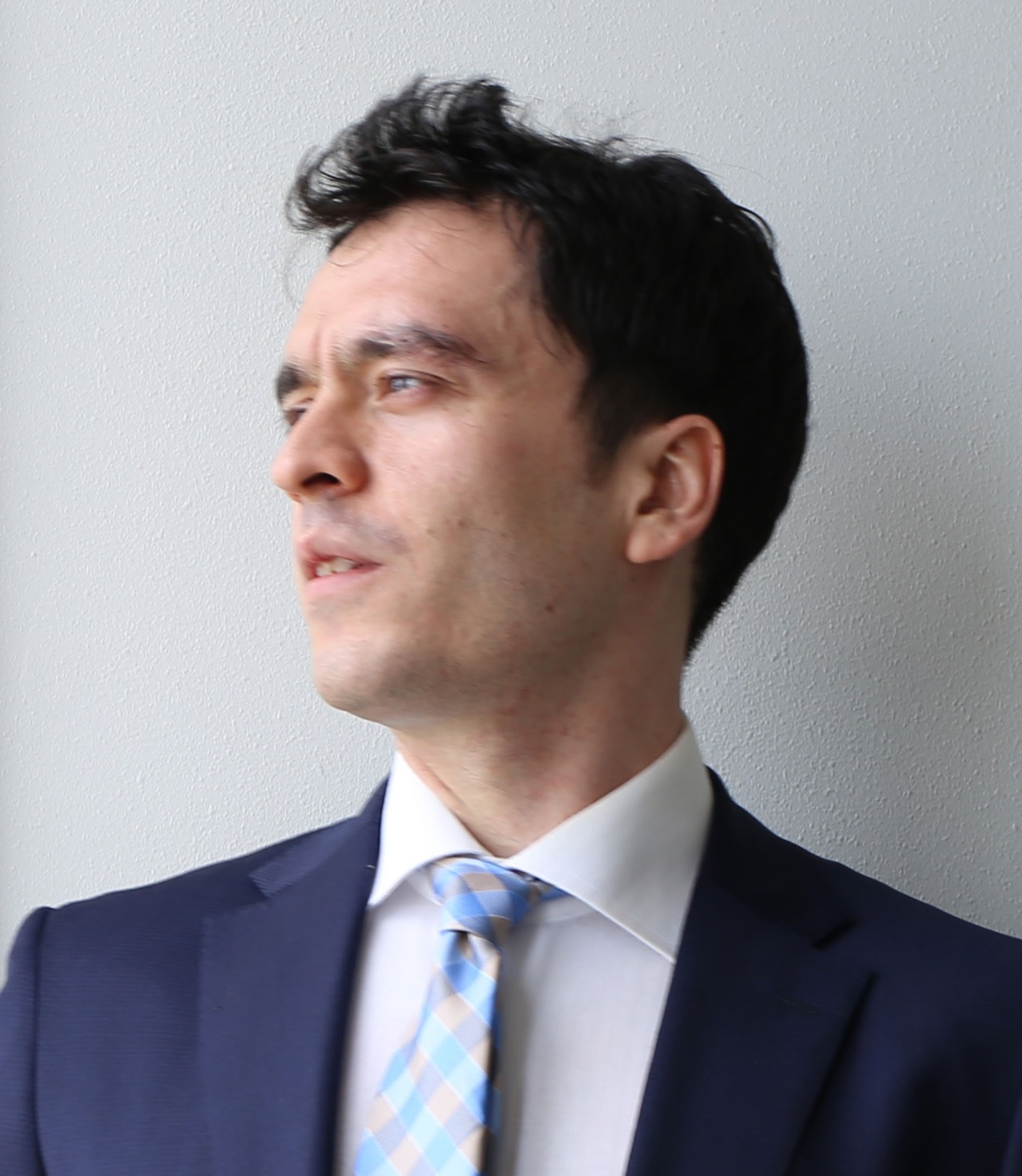}}]{Ekim Yurtsever} (Member, IEEE)
received his B.S. and M.S. degrees from Istanbul Technical University in 2012 and 2014 respectively. He received his Ph.D. in Information Science in 2019 from Nagoya University, Japan. Since 2019, he has been with the Department of Electrical and Computer Engineering, The Ohio State University as a research associate. 
	
His research focuses on artificial intelligence, machine learning, computer vision, reinforcement learning, intelligent transportation systems, and automated driving systems.

\vspace{-0.7cm}
	
\end{IEEEbiography}

\begin{IEEEbiography}[{\includegraphics[width=1in,height=1.25in,clip,keepaspectratio]{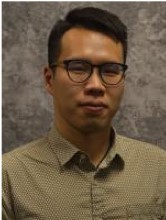}}]{Dongfang Yang} received his bachelor's degree in microelectronics from Sun Yat-sen University, Guangzhou, China, in 2014. He had been with The Ohio State University since 2015 and received his Ph.D. in Electrical and Computer Engineering from The Ohio State University, Columbus, OH, United States., in 2020. He is currently a senior algorithm engineer at Changan Automobile and a postdoc researcher at Chongqing University in Chongqing, China. His research interests include data analysis, machine learning, deep learning, and control systems, with applications in behavior prediction, decision-making, and motion planning in autonomous systems.
\end{IEEEbiography}


\begin{IEEEbiography}[{\includegraphics[width=1in,height=1.25in,clip,keepaspectratio]{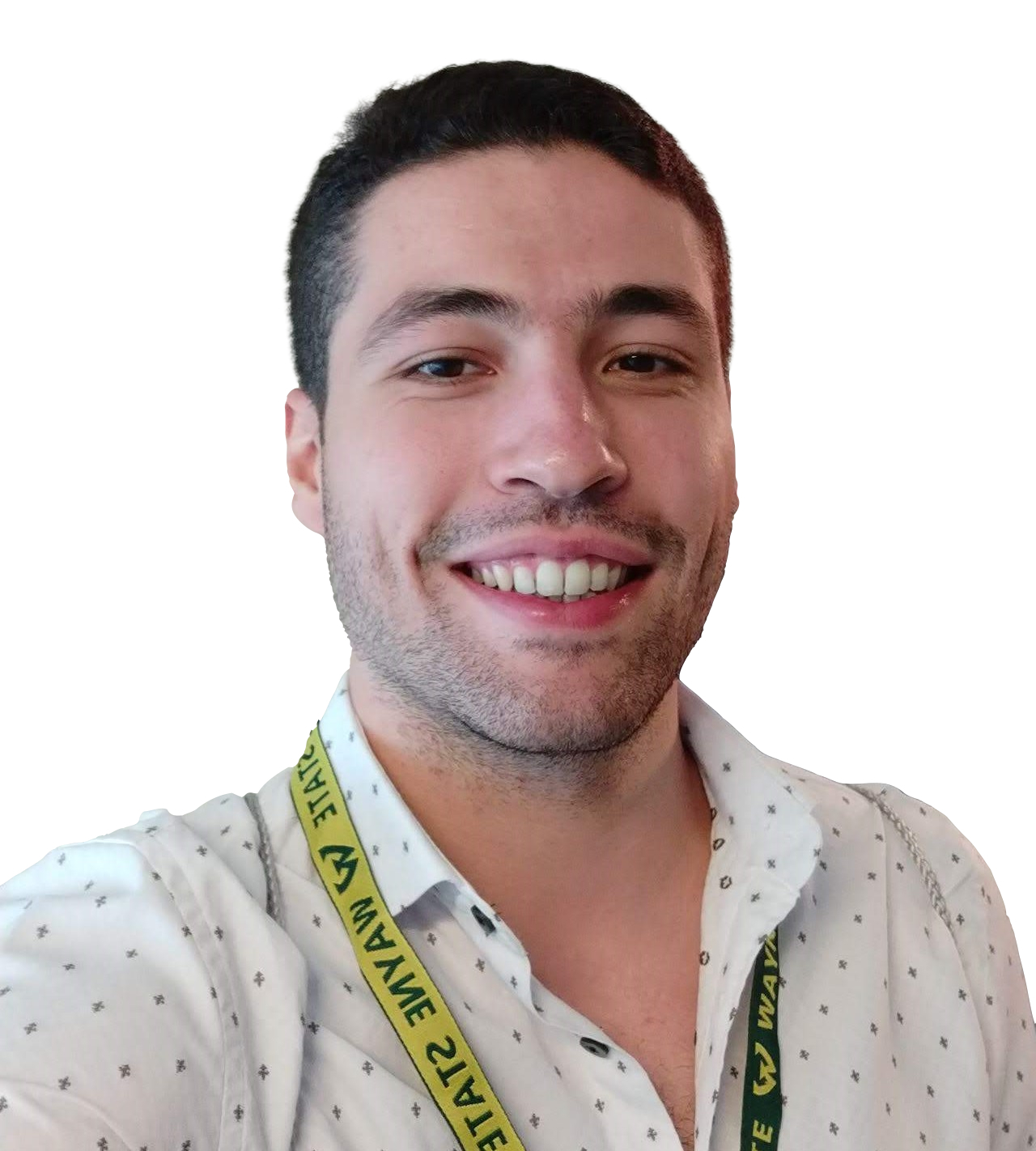}}]{Mert Koc} received his bachelor's degree in Electrical and Electronics Engineering from Middle East Technical University (METU), Ankara, Turkey, in 2018. He has been with The Ohio
	State University since 2018 and is going to receive his MSc. in Electrical and
	Computer Engineering from The Ohio State University in 2021. He is
	currently a graduate research associate at The Ohio State University.
	His research interests include computer vision, robotics and
	machine learning with applications in autonomous driving.\end{IEEEbiography}

\begin{IEEEbiography}[{\includegraphics[width=1in,height=1.25in,clip,keepaspectratio]{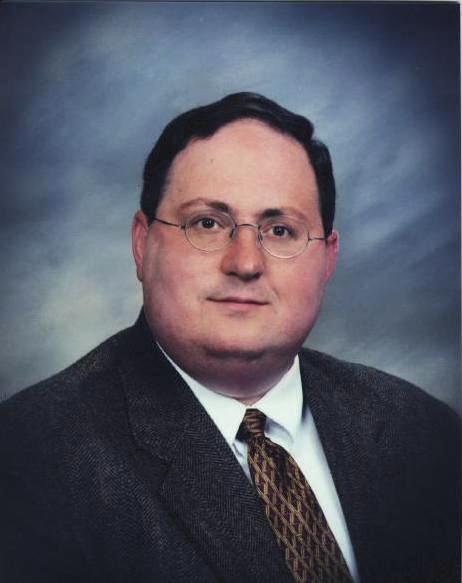}}]{Keith A. Redmill}
	(S'89–M’98–SM’11) received the B.S.E.E. and B.A. degrees in mathematics from Duke University, Durham, NC, USA, in 1989 and the M.S. and Ph.D. degrees from The Ohio State University, Columbus, OH, USA, in 1991 and 1998, respectively. Since 1998, he has been with the Department of Electrical and Computer Engineering, The Ohio State University, initially as a Research Scientist. He is currently a Research Associate Professor. He is a coauthor of the book \textit{Autonomous Ground Vehicles}. 
	He has significant experience and expertise in intelligent transportation systems, intelligent vehicle control and safety systems, sensors and sensor fusion, wireless vehicle to vehicle communication, multi-agent systems including autonomous ground and aerial vehicles and robots, systems and control theory, virtual environment and dynamical systems modeling and simulator development, traffic monitoring and data collection, and real-time embedded and electromechanical systems.
\end{IEEEbiography}










\end{document}